\documentclass[runningheads]{llncs}
\usepackage{amsmath,graphicx}
\usepackage{amsfonts,amssymb}
\usepackage{changepage}
\usepackage{multirow}
\usepackage{xcolor}
\usepackage{soul}
\usepackage{appendix}
\usepackage{color}
\usepackage{xspace}
\usepackage{stmaryrd}
\usepackage{array}
\usepackage{ mathrsfs }

\def\mat#1{\mathchoice{\mbox{\boldmath$\displaystyle\tt#1$}}
{\mbox{\boldmath$\textstyle\tt#1$}}
{\mbox{\boldmath$\scriptstyle\tt#1$}}
{\mbox{\boldmath$\scriptscriptstyle\tt#1$}}}

\def\vec#1{\mathchoice{\mbox{\boldmath  $\displaystyle\bf#1$}}
{\mbox{\boldmath  $\textstyle\bf#1$}}
{\mbox{\boldmath  $\scriptstyle\bf#1$}}
{\mbox{\boldmath  $\scriptscriptstyle\bf#1$}}}

\newlength{\colwidth}
\newcommand{\minus}{\scalebox{0.5}[1.0]{$-$}}

\newcommand{\node}{\vartheta \xspace} 

\newcommand{\Nf}{F\xspace} 
\newcommand{\No}{O\xspace} 

\begin{document}

\title{Visual Graphs from Motion (VGfM): Scene understanding with object geometry reasoning } 
\titlerunning{Visual Graphs from Motion} 


\author{Paul Gay \and
James Stuart \and Alessio {Del  Bue}}
%

\authorrunning{P. Gay et al.} 


\institute{Visual Geometry and Modelling (VGM) Lab,
Istituto Italiano di Tecnologia (IIT)\\
Via Morego 30, 16163 Genova, Italy\\
\email{\tt\small \{paul.gay, stuart.james, alessio.delbue\}@iit.it}
}

\maketitle

\begin{abstract}
Recent approaches on visual scene understanding attempt to build a scene graph -- a computational representation of objects and their pairwise relationships. Such rich semantic representation is very appealing, yet difficult to obtain from a single image, especially when considering complex spatial arrangements in the scene.
Differently, an image sequence conveys useful information using the multi-view geometric relations arising from camera motions. Indeed, object relationships are naturally related to the 3D scene structure.
To this end, this paper proposes a system that first computes the geometrical location of objects in a generic scene and then efficiently constructs scene graphs from video by embedding such geometrical reasoning. 
Such compelling representation is obtained using a  new model where geometric and visual features are merged using an RNN framework. We report results on a dataset we created for the task of 3D scene graph generation in multiple views.
\end{abstract}

\section{Introduction}

\begin{figure}[tp]
	\centering
	\includegraphics[width=\linewidth]{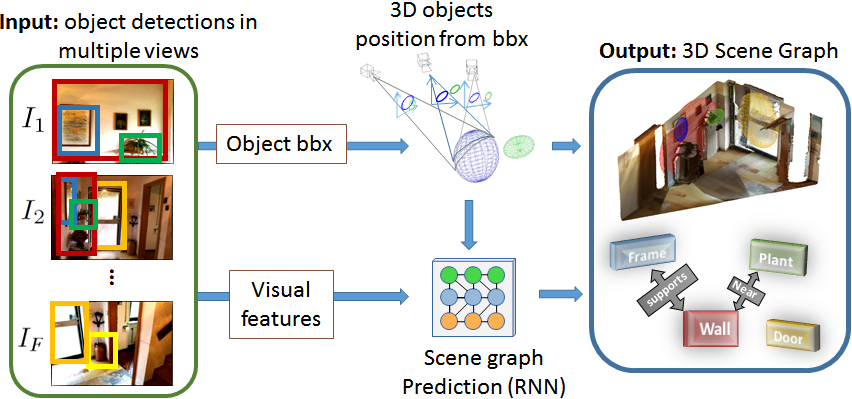} 
	\caption{Overview of the Visual  Graphs from Motion (VGfM) approach for 3D scene graph generation from multiple views. As input, an object detector extracts and matches 2D bounding boxes from objects in multiple images. The 3D position and occupancy of each object are estimated and, in parallel, visual features are extracted from each bounding box. These elements are then used to predict a 3D graph where each edge defines the semantic relationship between a pair of ellipsoids}
	\label{fig:goal}
\end{figure}

The ability to automatically generate semantic relationships between objects in a scene is useful in numerous fields. As such, in recent years there has been a significant amount of research toward this goal~\cite{lu2016visual,peyre2017weakly,zhang2017visual,dai2017detecting,liao2017natural} leading to the proposal of encoding  relationships  using a scene graph \cite{johnson2015image}. 




Common approaches for constructing scene graphs utilize visual appearance to guide the process, relying mainly on extracted Convolutional Neural Network (CNN) features. However, CNN visual features  fail to encode spatial relationships due to their invariance properties~\cite{sabour2017capsnet}. This is further compounded when considering complex 3D scenes where relationship predicates can become ambiguous and not easily solvable from a single view. This is exemplified in Fig.~\ref{fig:goal}, considering the image $I_{2}$ the plant could be, `\textit{near} the wall' or `\textit{supported by} the wall'. This ambiguity can be rectified through the understanding provided by adjacent images, resolving for the camera pose and predicting the \textit{near} predicate -- `plant \textit{near} the wall', as the \textit{support} predicate would be related to the shelf. In this paper we aim to encode the required information using the knowledge of the 3D geometry of the objects in the scene. 


We therefore propose to combine the advantages of both visual and geometric information to efficiently predict spatial relations between objects as shown in Fig.~\ref{fig:goal}. Given a set of 2D object bounding box detections matched across a sequence of images, using multi-view relations we compute the 3D locations and occupancies of objects described as a set of 3D ellipsoids. 
At the same time, we extract visual features from each object detection to model their visual appearance. These two representations are given as input to a Recurrent Neural Network (RNN) which has learned to predict a coherent scene graph where the objects are vertices and their relationships are edges.  
Overall, our Visual Graphs from Motion (VGfM) approach is appealing as it combines geometric and semantic understanding of an image, which has been a long term goal in computer vision~\cite{bao:etal:2012,Sengupta:2013,reddy2015dynamic,sung2015data,choy20163d,factored3dTulsiani17,Hane:2017}. 
We demonstrate the effectiveness of such a representation by creating a new dataset for 3D scene graph evaluation that is derived from the data provided in ScanNet~\cite{dai2017scannet}.
To summarize, our contributions in this paper are:
\begin{itemize}
	\item To define the problem and the model related to the computation of 3D scene graph representations across multiple views; 
	\item To extract reliable geometric information in multiple views, we propose an improved geometric method able to estimate objects position and occupancy in 3D, modelled as a set of quadrics;  
	
	\item Finally, to provide a new real world dataset, built over ScanNet, which can be used to learn and evaluate 3D scene graph generation in multiple views~\footnote{Code and data can be found at: \url{https://github.com/paulgay/VGfM}}.
\end{itemize}

The paper is structured as follows. Relevant literature to the VGfM approach is reviewed in Sec.~\ref{sec:relwork}. We outline our refined strategy for object 3D position and occupancy in Sec.~\ref{sec:quadrics}, then present VGfM and its learning procedure in Sec.~\ref{sec:themodel}. The dataset is described in Sec.~\ref{sec:dataset} with detailed evaluation of VGfM performance and the benefit of geometry refinement. We then conclude the paper in Sec.~\ref{sec:conc}.

\section{Related work}
\label{sec:relwork}
We now review the 3 topics related to our approach: scene graph generation from images, classification from videos and 3D object occupancy estimation.

Early works on visual relation detection  were training classifiers to detect each relation in an image independently from each other~\cite{desai2010discriminative,lu2016visual}. However, a scene graph often contains  chains of relationships for instance: \textit{A man HOLDING a hand BELONGING TO a girl}. Intuitively, a model able to leverage on this fact should obtain more coherent scene graphs. To account for this, Xu \textit{et al.} ~\cite{xu2017scene} proposed a model which explicitly defines a 2D scene graph. 
The framework naturally deals with chains of relations because inference is performed globally over all the objects and their potential relations. To this end, a message passing framework was developed using standard RNNs. 
This is in line with current approaches which combine the strengths of graphical models and neural networks~\cite{liang2016semantic,zheng2015conditional}. Each object and relation represents a node in a two layer graph and is modelled by the hidden state of an RNN. The state of each node is refined by the messages sent from adjacent nodes. 
This architecture has the flexibility of graphical models and thus can be used to merge heterogeneous sources of information such as text and images~\cite{li2017scene}. We utilize this mechanism in our model while extending it to incorporate the 3D geometry. To the best of our knowledge, this is the first time that geometric reasoning is exploited for scene graph generation.
%

Object detection within a sequence (video) is largely still reliant on temporal confidence aggregation across image detections or applying RNN for temporal memory \cite{tripathi2016objrnnvid}. With the difficulty of predicting confidence within a CNN \cite{nguyen2015deep} these approaches rely on detection consistency. 
Alternatively, more advanced video tubelets in T-CNN~\cite{kang2016object} are optimized for the detection confidence. In a similar way, we exploit the multiple view information within our model by including a fusion mechanism based on message passing across images.

Recently, new techniques have emerged to estimate the 3D spatial layout of the objects as well as their occupancy~\cite{rubino20173d,dong2017visual,cvpr17chen}. These techniques rely on the quality of deep learning object detectors \cite{rubino20173d,dong2017visual} or the use of additional range data~\cite{cvpr17chen}. Similarly volumetric approaches have been used to construct the layout of objects in rooms, or construct objects and regress their positioning~\cite{factored3dTulsiani17}. These strategies provide alternative representations for scene graph generation since they associate object labels to the 3D structure of the scene, but lack the relationships required to construct a scene graph. 
In particular the approach localization from Detection (LfD)~\cite{rubino20173d} 
leverages 2D object detector information to  obtain the 3D position and the occupancy of a set of objects represented through quadrics. Although ellipsoids are an approximation of the region occupied by an object, they provide the necessary support for spatial reasoning in a closed form which can be efficiently computed. However, in the current methods~\cite{rubino20173d,gay2017probabilistic}, there is no explicit constraint to enforce the quadric to be a valid ellipsoid. As a consequence, low baselines and inaccurate bounding boxes might result in degenerate quadrics. 
In the next section, we present an extension named LfD with Constraints (LfDC) which is based on linear constraints on the quadric centers. It has the advantage of being a fast closed-form solution 
while being more robust than LfD~\cite{rubino20173d}.

\section{Robust object representation with 3D quadrics}
\label{sec:quadrics}
Even if they are an approximate representation of objects, a representation based on ellipsoids (or formally quadrics) can be embedded in the graph effectively with multiple views (as described in Sec.~\ref{sec:themodel}). In this section, we briefly consider the prior work for generating quadrics from multi-view images, then resolve for their limitations so making the approach more suitable for scene graph construction.

Let us consider a set of image frames $f=\{1 \dots \Nf\}$  representing a 3D scene under different viewpoints. A set of $i=\{1 \dots N\}$ rigid objects is placed in arbitrary positions. We assume that each object is detected in at least $3$ images. Each object $i$ in each image frame $f$ is given by a $3 \times 3$ symmetric matrix $\mat C_{if}$ which represents an ellipse inscribed in the bounding box as shown in Fig. \ref{fig:goal} (left \& top middle). The aim is to estimate the $4 \times 4$ matrix $\mat Q_i$ representing the 3D ellipsoid  whose projection onto the image planes best fit the measured 2D ellipses $\mat C_{if}$. The relationship between $\mat{Q}_i$ and their reprojected conics $\mat{C}_{if}$ is defined by the $3\times4$ perspective camera matrices $\mat P_{f}$ which are assumed to be known (i.e. the camera is calibrated). 
%
The LfD method described in~\cite{rubino20173d} solves the problem in the dual space where it can be linearized as:  
\begin{equation} \label{eq:eqOut}
\beta_{if} {\vec c}_{if} = \mat G_{f} \vec v_{i},
\end{equation}
where $\beta_{if}$ is a scaling factor, the $6$-vector ${\vec c}_{if}$ is the vectorised conic of the object $i$ in image $f$, the $10$-vector $\vec v_{i}$ is the vectorised quadric and the matrix $\mat G_f$ contains the elements of the camera projection matrix after linearization\footnote{the supplemental material provides more mathematical details about this step}.
Then, stacking column-wise Eq. (\ref{eq:eqOut}) for $f = 1 \dots F$, with $F \geq 3$, we obtain:

\begin{equation}\label{eq:eqsys}
\mat M_{i} \vec w_{i} = \vec 0_{6F},
\end{equation}
where $\vec 0_{x}$ denotes a column vector of zeros of length $x$, and the matrix $\mat M_i \in \mathbb{R}^{6F \times (10+F)}$ and the vector $\vec w_i \in \mathbb{R}^{10+F}$ are defined as follow:
\begin{equation}\label{eq:eqsysmx}
\mat M_{i} = 
\arraycolsep=1.4pt
\begin{bmatrix}
\mat G_1 	& \minus {\vec c}_{i1}	& \vec 0_6 		& \vec 0_6 		& \ldots & \vec 0_6 \\
\mat G_2 	& \vec 0_2   	& \minus {\vec c}_{i2}	& \vec 0_2  		& \ldots & \vec 0_2 \\
\vdots 		&   	&  		&    		& \ddots &  \\
\mat G_F	& \vec 0_2  	& \vec 0_2 		& \vec 0_2   		& \ldots & \minus {\vec c}_{iF} \\
\end{bmatrix}, \;\;\;
\vec w_{i} = 
\begin{bmatrix}
\vec v_i\\
\vec{\beta}_i
\end{bmatrix},
\end{equation}
where $\vec{\beta}_i = \left[ \beta_{i1},\beta_{i2},\cdots,\beta_{iF} \right]^{\top}$ contains the scale factors of the object $i$ for the different frames.

Since the object detector can be inaccurate, it makes sense to find the quadric $\tilde{\vec{w}}_{i}$ by solving the following minimization problem:
\begin{equation} \label{eq:argmin}
 \tilde{\vec{w}}_{i}= \arg \min_{\vec w} \| \mat M_i \vec w \|_2^2,~~~  s.t. \|\vec w\|^2_2=1,
\end{equation}
where the equality constraint $\|\vec w\|^2_2=1$ avoids the trivial zero solution. 
The solution of this problem consists in computing the $\mat{SVD}$ on the $\mat M_i$ matrix and taking the right singular vector associated to the minimum singular value.

\begin{figure}[t!]  	
	\centering
	\includegraphics[width=\linewidth]{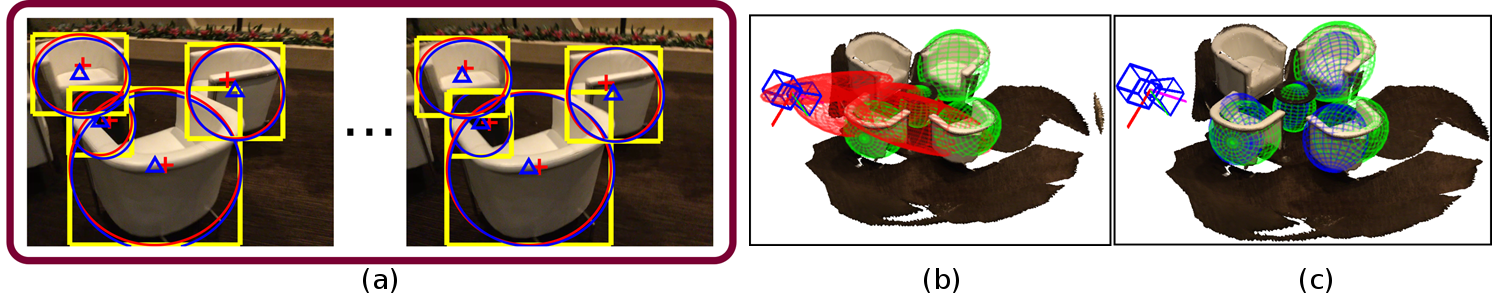}
	\caption{(a) are example ellipse reprojections of LfD~\cite{rubino20173d} in red and LfDC in blue with cross and triangle respectively for the centers. (b, c) is the point cloud, 3D quadrics (using same color labelling), camera poses for two camera, the ground truth is shown in green.  It can be seen that the proposed solution overcomes the limitation of~\cite{rubino20173d}.}
	
	\label{fig:scan}
\end{figure}

However, the algebraic minimization in Eq.~(\ref{eq:argmin}) does not enforce the obtained quadric to be a valid ellipsoid. As can be seen in Fig.~\ref{fig:scan}, fitted ellipsoids can be inaccurate despite giving a reasonable 2D projection. The proposed LfDC solution generates ellipsoids as Fig.~\ref{fig:scan}c, and in turn improves overall performance.

A common indication of the estimated quadric being degenerate can be fairly guessed by checking where the estimated ellipse center is located. If the center is outside the boundaries of the estimated ellipse contour, this clearly points out to a degenerate configuration. Given this last observation, rather than constraining directly the solution to lie in a valid ellipsoid parameter space, we include a set of equations imposing the reprojection of the center of the 3D ellipsoid being closer to the centers of the ellipses. 
%
This can be done by adding an additional set of rows in the matrix $\tilde{\mat M}_i$ used in Eq.~(\ref{eq:eqsys}).


%


This constraint can be added by observing that  the center parameters of the vectorised dual quadric $\vec{v}$ appear separately in linear terms\footnote{We refer to supplemental material for further mathematical details.} at position 4, 7 and 9 in the vector. The same fact holds for the vectorised conic $\vec{c}$ at positions 3 and 5 (we omit indexes to simplify the notation):
\begin{equation}
\label{eq:conic}
\vec c^* = \vec c_{3,5} =  
\left[
\begin {array}{cc}
 -t^c_1 &  -t^c_2 
\end {array} 
\right], \: \: \:
\vec v^* = \vec v_{4,7,9} = 
\left[ \begin {array}{ccc} 
-t_{{1}} & -t_{{2}} & -t_3 
  \end {array}
\right],
\end{equation}
where $\vec c^*$ and $\vec v^*$ contain the centers of the ellipse and the ellipsoid respectively. 
We can use this fact to directly include the equations which enforce the ellipsoid center to be projected in the centers of the ellipses. Given a frame $f$ and an object $i$, the constrained equations are:
\begin{equation}
\mat G^c_f \vec  v_i^* = \vec c^*_{if} \beta_{if}, 
\end{equation}
with the $2 \times 10$ matrix $\mat G^c_f$ defined as: 
\begin{equation} 
		\arraycolsep=1.7pt
		\mat G^c_f \scalebox{0.5}[1.0]{$=$}
		\left[ \begin {array}{cccccccccc} 0 & 0 & 0 & p_{{11}} & 0 & 0 &p_{{12}} &0 & p_{{13}}& p_{{14}}\\ 
		\noalign{\medskip}0 & 0 & 0 & p_{{21}} & 0 & 0 &p_{{22}} &0 & p_{{23}}& p_{{24}}\end {array} \right],
\end{equation}		
where each value $p_{ij}$ corresponds to an element of the camera matrix $\mat P_f$.
These equations are included in the system of Eq.~(\ref{eq:eqsys}) by replacing the matrix $\mat M_{i}$ by $\tilde{\mat M}_{i}$ such that:
\begin{equation}\label{eq:eqsys2}
\tilde{\mat M}_{i} \vec w_{i} = \vec 0_{8F},
\end{equation}
where the matrix $\tilde{\mat M}_i \in \mathbb{R}^{8F \times (10+F)}$ is defined as follow:
\begin{equation}\label{eq:eqsysmx_cen}
\tilde{\mat M}_{i} = 
\arraycolsep=1.4pt
\begin{bmatrix}
\mat G_1 	& \minus {\vec c}_{i1} 	& \vec 0_6 		& \vec 0_6 		& \ldots & \vec 0_6 \\
\mat G^c_1 	& \minus {\vec c}_{i1}^* 	& \vec 0_6 		& \vec 0_6 		& \ldots & \vec 0_6 \\
\vdots 		&   	&  		&    		& \ddots &  \\
\mat G_F	& \vec 0_6  	& \vec 0_6 		& \vec 0_6   		& \ldots & \minus {\vec c}_{iF} \\
\mat G^c_F	& \vec 0_6  	& \vec 0_6 		& \vec 0_6   		& \ldots & \minus c^*_{iF} \\
\end{bmatrix}.
\end{equation}
The solution of this new system can then be obtained with the SVD of the $\tilde{\mat M}_{i}$ matrix as done for the minimization problem described in Eq.~(\ref{eq:argmin}).
This method, named LfDC, has both the effect of reducing the number of degenerated quadrics (i.e. to localize more objects in the scene) and to improve the quality of object localizations and occupancy estimation as it will be shown in the experimental section. For these reasons, LfDC also enables to improve the performances when estimating the scene graphs using multi-view relations. 

\section{Scene graphs from multiple images} \label{sec:themodel}
\begin{figure*}[t]  	
	\centering
	\includegraphics[width=\linewidth]{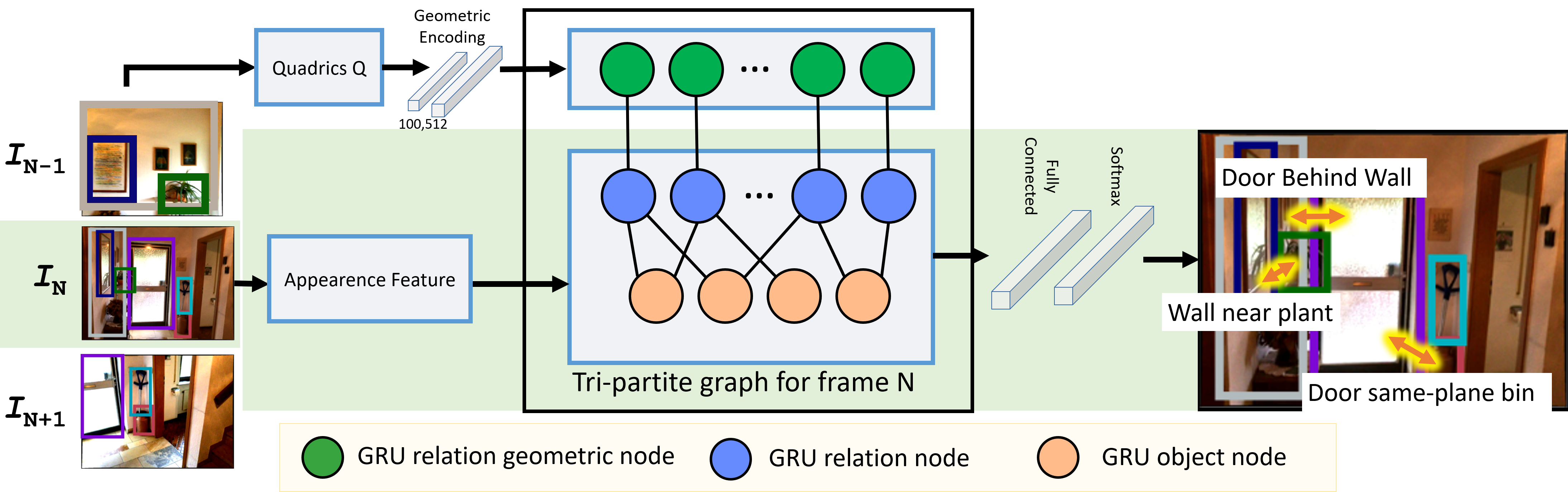} 
	\caption{Our scene graph generation algorithm takes as input a sequence of images with a set of object proposals (as ellipsoids). 
	In addition, visual features are extracted for each of the bounding boxes, these features are then fed to initialise the GRU object and relation nodes. A tri-partite graph connects the object, relation and geometric nodes and iterative message passing updates the hidden states of the object and relation nodes. At the conclusion of the message passing the scene graph is predicted by the network and then the next image of the sequence is  processed.}
	\label{fig:method}
\end{figure*}

The VGfM approach models the scene graph within a tri-partite graph which takes as input the features, both visual and geometric (from Sec.~\ref{sec:quadrics}), and outputs the prediction of the object labels and predicates, as illustrated on Fig~\ref{fig:method}. 
The graph merges geometric and visual information, as well as refining jointly the state of all the objects and their relationships.
This process is performed iteratively over each of the $\Nf$ images of the sequence. 

Therefore, let $G=(\node,E)$ denotes the tri-partite graph of a current image. We define $\node$ as the set of nodes that corresponds to attributes, defined as $\node=\{\node^g,\node^o,\node^r\}$ related to geometry, objects and relationships respectively, while $E$ refers to pairwise edges which connect each object with its relation. 
The set of object nodes is denoted as $\node^o=\{\node^o_{i},i=1\dots \No \}$ and models their semantic states. Similarly, $\node^r$ models the semantic states of the relationships and is defined as $\node^r=\{\node^r_{i \shortrightarrow j}, \: i=1\dots \No, \: j=1\dots \No, \: i\neq j \}$. Finally, $\node^g=\{\node^g_{i \shortrightarrow j} , \: i=1\dots \No, \: j=1\dots \No, \: i\neq j \}$ is the set of geometric nodes constructed over the quadrics previously computed expressing the geometric state of each relation (see Sec.~\ref{sec:geomnodes} for construction). 

The states of the graph are then iteratively refined by message passing among the nodes, exchanging information about their respective hidden states (see Sec.~\ref{sec:tri}). The hidden states $h_{i \shortrightarrow j}$ (resp. $h_i$) of each relation node $\node^r_{i \shortrightarrow j}$ (and resp. object node $\node^o_i$) are modelled with Gated Recurrent Units~\cite{cho2014properties} (GRU). This allows each node to refine its state by exploiting incoming messages from its neighbors. Differently from the object and relation nodes, each geometric node $\node^g_{i \shortrightarrow j}$ is considered as an observation and its state $g_{i \shortrightarrow j}$ is fixed, this allows the reliability of the geometric information to be enforced. 
If the geometric nodes are removed from the graph, we obtain the framework of~\cite{xu2017scene}.

After $K$ iterations of message passing the hidden states from the object and relation nodes are used to compute the  classification decision, i.e. object and relation labels, as provided by the final fully connected layer. This layer takes as input the hidden state of a relation node and produces a distribution over the relation labels through a softmax, this step is  performed to compute the object labels as well. We treat predicate labels as in the multi-label scenario where a predicate is detected for a given relation if the softmax score is higher than the label indicating its absence. We further outline the training specifics of the model in Sec.~\ref{sec:learning}. With the creation of the scene graph the next image in the sequence is then processed.

As our goal is to share information between images, we can encourage sharing beyond object and relation nodes and pass messages between images within the sequence. This can be simply performed by connecting tri-partite graph nodes ${\node^r,\node^o}$ among images and this process is explained in Sec.~\ref{sec:multi}.
%
\subsection{Construction of the geometric nodes}
\label{sec:geomnodes}
As described in Sec.~\ref{sec:quadrics}, we obtain a set of ellipsoids $Q=\{\mat Q_{i},i=1\dots \No \}$
from the object detections. We then extract the 3D coordinates of the center of each quadric $\mat Q_i$ and the six points at the extremities of its main axis. 
Finally, the geometric encoder takes as input the coordinates extracted from the ellipsoids $\mat Q_i$ and $\mat Q_j$ in order to produce the state of the geometric node $\node^g_{i \shortrightarrow j}$. This encoder consists of a multi-layer perceptron with two fully connected layers of sizes $100,512$. These values were identified empirically to give enough capacity to the network to link both the quadric positions and occupancies and the given complexity of the final labels. 
We additionally experimented with a bag-of-word based encoding, proposed in~\cite{peyre2017weakly}, and found similar performances.

\subsection{Message passing between nodes}
\label{sec:tri}
The refinement of the hidden states is carried out via message passing. At each inference iteration, messages are sent along the graph edges. Each relation node $\node^r_{i \shortrightarrow j}$ is linked by undirected edges to the object state nodes $\node^o_{i}$, $\node^o_{j}$ and the corresponding geometric node $\node^g_{i \shortrightarrow j}$. We use the message pooling scheme proposed in~\cite{xu2017scene}. 

At each iteration, the node $\node^o_{i}$ receives the following message: 
\begin{equation}
m_i = \sum_{j:i\shortrightarrow j} \sigma(\vec a_1[h_i,h_{i \shortrightarrow j}])h_{i \shortrightarrow j}  + \sum_{j:j\shortrightarrow i} \sigma(\vec a_2 [h_i,h_{j \shortrightarrow i} ]) h_{j \shortrightarrow i},
\end{equation}
where $[,]$ denotes the concatenation operator, $\sigma$ is the sigmoid function, $\{j:i\shortrightarrow j\}$ is the set of all the relations where object $j$ is present at the right of the predicate, and the weights $\vec a_1$ and $\vec a_2$ are learned.
The relationship nodes are also updated, where each node $\node^r_{i \shortrightarrow j}$ receives the following message:
\begin{equation}
m_{{i \shortrightarrow j}} = \sigma(\vec b_1[h_i,h_{i \shortrightarrow j}])h_i + \sigma(\vec b_2[h_j,h_{i \shortrightarrow j}])h_j + \sigma(\vec b_3[g_{i \shortrightarrow j},h_{i \shortrightarrow j}])g_{i \shortrightarrow j},
\end{equation}
where $b_1$, $b_2$ and $b_3$ are learned parameters.

As with loopy belief propagation, this can be seen as an approximation of an exact global optimization, enabling the refinement of each hidden state based on its context. Conversely to a classic message passing scheme, the last inference decision on the label values is not performed within the tri-partite graph but by using a last fully connected layer. On average in our experiments, the inference time is 0.25 second per image on a Tesla K80.

\subsection{Sharing information among multiple images}
\label{sec:multi}
\begin{figure*}[htbp!]  	
	\centering
	\includegraphics[width=\linewidth]{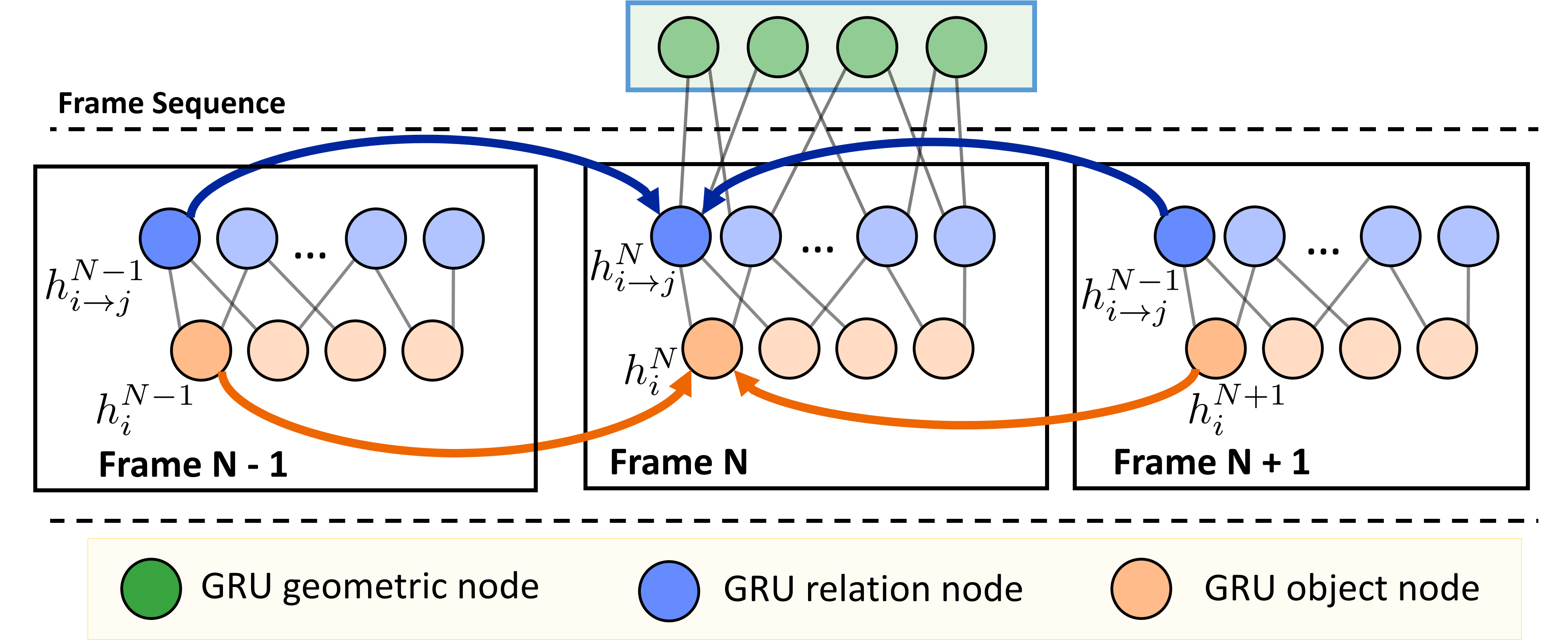} 
\caption{This figure displays how the graphs operating on a single frame (same as shown on Fig.~\ref{fig:method}) are linked by the fusion mechanism.}
\label{fig:fusion}
\end{figure*}

We now extend the proposed single image model to fuse information among the images of the sequence. In this case, the visual features can be shared where the network benefits from taking into account potential appearance changes as well as aiding consistency among the views.
To this end, we rely on the message passing mechanism and include cross-image links which connect the tri-partite graphs for each image. As shown in Fig.~\ref{fig:fusion}, 
each relation node receives messages from all the nodes modelling the same relation in the other images. The same principle is applied for the object nodes. 

We extend the notation so that it refers to nodes and messages image by image. Let us denote by $m^f_{{i \shortrightarrow j}}$ (resp. $m_i^f$) the message that the relation node $h^f_{i \shortrightarrow j}$ (resp. the object node $h_i^f$) appearing in the image $f$ receives from the other images. Then, we compute the messages with the following equations:

\begin{equation}
m^f_{{i \shortrightarrow j}} = \sum_{l,f\neq l} \sigma(\vec c_1[h^f_{i \shortrightarrow j},h^l_{i \shortrightarrow j}])h^l_{i \shortrightarrow j},
\end{equation}
\begin{equation}
m^f_i = \frac{1}{\Nf} \sum_{l,f\neq l} \sigma(\vec c_2[h^f_i,h^l_i])h^l_i,
\end{equation}
where $\vec c_1$ and $\vec c_2$ are learned weights. This formulation can be seen as weighted average of the visual features were the weights are learned as an attention mechanism. This new cross-image message is then added to the local one described in Sec.~\ref{sec:multi} to form the final message.


\subsection{Learning}\label{sec:learning}
Our model is trained with cross-entropy loss. 
We also use similar hyper-parameters to Xu et al.~\cite{xu2017scene} with a learning rate of  $1e^{-3}$ and $K=2$ iterations of message passing. Batches of 8 images were used for the single image system. For the multi-image approach described in Sec.~\ref{sec:multi}, each batch corresponds to one image sequence. We reduce the sequence to $10$ images selected uniformly to save memory space. In contrast to~\cite{xu2017scene}, we retain all region proposals as we are considering the ellipsoid proposal that already prunes the per-frame object proposals. 
We extract visual features from VGG-16~\cite{simonyan2014very} pretrained on MS-COCO and use the $FC_{7}$ layer to initialize the hidden states of the RNNs.
The RNNs are trained while keeping the weights of the visual features fixed. Two sets of shared weights are optimized during training: one for the objects  and one for the relations. 
The state of the GRU for both input and output has a dimension of $512$.

\section{Dataset description and experimental evaluation}\label{sec:dataset}
%

%
Prior datasets for the scene graph generation problem are based on singular images with relationship annotations, but in general they do not have multi-view image sequences necessary to exploit the proposed model.
We thus create GraphScanNet by manually extending and upgrading the ScanNet dataset~\cite{dai2017scannet} with relationships between the annotated objects. The ScanNet dataset provides $2.5$ million views in more than $1500$ scans annotated with semantic and instance level segmentation. 3D camera poses are also provided as estimated from an online 3D reconstruction pipeline (BundleFusion~\cite{dai2017bundlefusion}) algorithm run on the RGB-D images. Since VGfM does not require depth, we also tried a visual SLAM algorithm~\cite{mur2015orb}, but we found that the results were not accurate enough.

Although one thousand object categories are present, we refine the list of objects to resolve for annotator errors and the frequency of object occurrences in sequences resulting in a refined list of $34$ object categories.
Our annotations are a set of $8762$ view-independent compositional relationships between couples of 3D objects. Our proposed predicates are inspired by Visual Genome~\cite{krishna2016visualgenome}, but we opt for a concise set that is loosely aimed to encompass many relationships that can occur within the sequences. It can be seen from the ScanNet class labels that when annotators are given expressive freedom in labels, cultural or personal bias can make annotations implausible for learning systems where many objects are synonyms or localized vernaculars.
Our predicates are as follows:
\begin{description}
	\setlength\itemsep{0em}
	\item \textit{Part-of:} A portion or division of a whole that is separate or distinct; piece, fragment, fraction, or section 
	, e.g. shelf is \textit{part-of} a bookcase. 
	\item \textit{Support:} To bear or hold up (a load, mass, structure, part, etc.); serve as a foundation for
	. Where hypernyms could be considered support from \textit{behind, below, hidden}; in our case `below' is most prevalent. 
	\item \textit{Same-plane:} Belonging to the same or near similar vertical plane in regards to the ground normal. As an example, a table might be \textit{same-plan} as a chair.
	\item \textit{Same-set:} Belonging to a group with similar properties of function. The objects could  define a region, e.g. in Fig.~\ref{fig:qual} the table, chair and plate belong to the same set whereas the shoes on the floor are separated. This is similar to the concept of scenario recently studied in~\cite{daniels2018scenarios}, and where they proved this being a powerful clue for scene understanding.
\end{description}

As relationships are derived from images, there are differences in terms of number of instances for each predicate. \textit{Same-set} and \textit{Same-plane} appear about $30,000$ times in the images, whereas \textit{Support} $3,000$ times and \textit{Part-of} only $600$ times. This has an impact on the performances as explained in the evaluation.

The 3D object segmentation enables us to construct a 3D ground-truth (GT) by fitting ellipsoids to each object mesh. Object bounding box in 2D are also computed by projecting each object point cloud into the image. Such  bounding boxes are created by fitting a rectangle that encloses the set of  2D points. We then automatically extracted 2000 sequences coming from 700 different rooms with at least 4 objects in each of them. 
These sequences are challenging for 3D reconstruction, since the recording of the rooms was done by rotating the camera with limited translation motions. On average, the angle spanned by the camera trajectory is $4.3$ degrees. 
\subsection{Evaluation of the quadric estimation}
\begin{figure}[t]
    \centering
        \begin{tabular}{ccc}
         \includegraphics[width=0.32\linewidth]{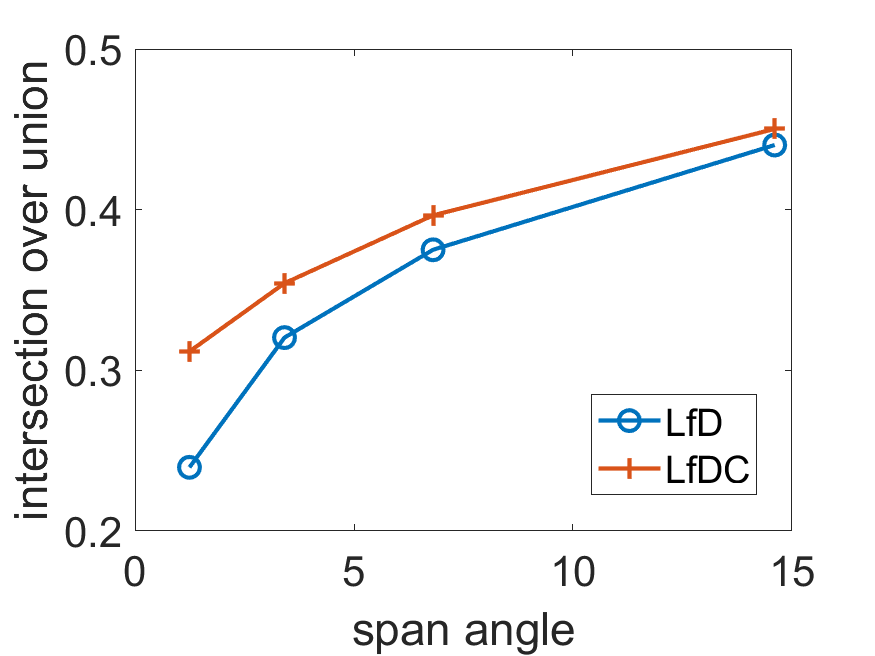}&\includegraphics[width=0.32\linewidth]{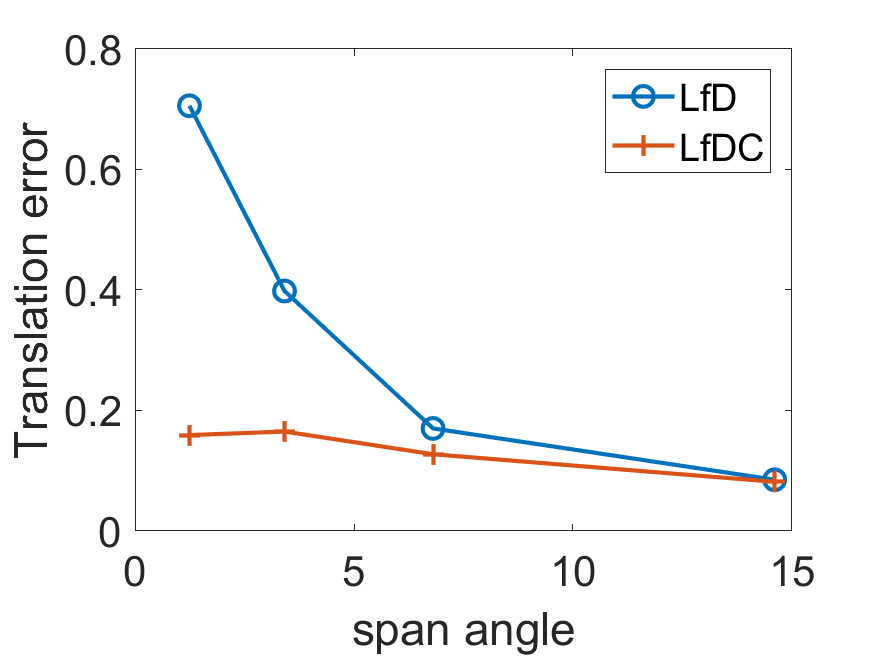} & \includegraphics[width=0.32\linewidth]{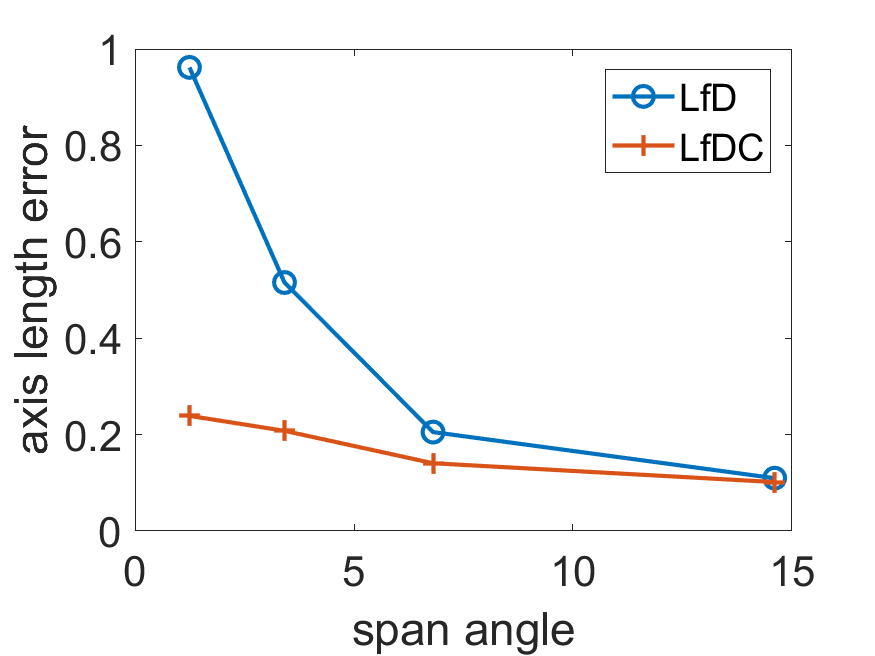}\\
    \end{tabular}
    \caption{Evaluation in terms of $O_{3D}$ accuracy, translation and axes length errors for the LfD~\cite{rubino20173d} method and our proposed approach LfDC.}
    \label{fig:res}
\end{figure}
We first evaluate how accurate are the quadrics obtained from the different methods.
We run the original LfD method~\cite{rubino20173d} on the extracted sequences and compared with the ones from our LfDC approach. 
On the $1979$ sequences, we measured that only 48\% of the quadrics estimated by LfD are valid ellipsoids. This number rises up to  60\% when we use our LfDC method. This validates our initial hypothesis that the additional equations are useful to avoid non-valid quadrics. 
In the following, we evaluate the accuracy of the ellipsoids by considering only the ones who are found valid by all methods.

One of the main limitations of LfD is the sensibility when the image sequences have a short  baseline (i.e. short camera path and/or very few image frames). 
To study this effect, the error and accuracy values are plotted in function of the maximum angle spanned by the camera during the sequence where the object is recorded. In  Fig.~\ref{fig:res}, we compare the methods according to three metrics: $O_{3D}$, which is the intersection over union between the proposed and the GT quadrics, the translation and the axis length errors.


We can see that the LfDC outperforms the previous LfD method in terms of the three metrics: volume overlap, translation and axis length error. The constraints on the centers are beneficial to improve on these three aspects since the solution is still computed globally for all the quadric parameters. Secondly, we observe that, although relatively small in average, the improvements are important in case of a low baseline. 



\subsection{Evaluation on the scene graph classification task}
We evaluate our systems on the tasks of object and relation classification i.e. given the bounding boxes of a pair of objects, the system must predict both the object classes and the predicate of their relation. These two tasks encompass the problem of scene graph generation when performed recursively over the image where annotations are performed in terms of multi-label fashion i.e. presence and absence. We selected 400 rooms for training, 150 as a validation set and 150 for testing.

We first study the influence of the quadric estimation algorithms. We run our VGfM and use as input the ellipsoids provided by LfD, LfDC and GT quadrics. Results are reported in Table~\ref{tab:quadrics}. 
\begin{table}[tbh]
    \renewcommand{\arraystretch}{1.3}
	\centering
	\begin{tabular}{l|m{1.5cm}||m{1.5cm}|m{1.5cm}}
               		&    GT &  LfD\cite{rubino20173d}       & LfDC  \\
		\hline
		Object label & 76\% &	 75\%              & 75\% \\
		\hline
		\hline
		Same-plane   & 75\% &   72\%                       & \textbf{74}\% \\
		Same-set     & 62\%&   59\%              & \textbf{61}\%  \\
		Support      & 69\%&   64\%                       & \textbf{67}\% \\
		Part-of      & 69\%&   65\%                       & \textbf{69}\% \\ 
		\hline
	\end{tabular}
	\caption{Comparison of the use of different quadrics to classify the scene graphs. The numbers in bold are related to the best results LfD and LfDC.}
	\label{tab:quadrics}
\end{table}
We can see that the differences between the different methods are relatively small, but still coherent with the accuracy reported in Fig.~\ref{fig:res}. The LfD obtains the worst results and the best performing method is LfDC. Overall, the use of GT quadrics brings an additional improvement, but the accuracy remains relatively close to the other methods.

We now study the influence of the different components of the system in an ablation study in Table~\ref{tab:predicate}. The baseline \cite{xu2017scene} uses only visual appearance. The method VGfM-2D corresponds to a variation of our method without 3D information where we computed the geometric states from the coordinates of the 2D bounding boxes instead of using the ellipsoids. To evaluate the potential of using geometry alone, we also report results while using only the geometric encoder described in Sec.~\ref{sec:geomnodes}. A softmax layer is appended to this encoder in order to use it as classifier. The resulting network is then trained from scratch for the tasks of predicting predicates and object labels. VGfM + Fusion corresponds to the addition of the fusion mechanism over multiple images described in Sec.~\ref{sec:multi}.
\begin{table}[bth]
	\centering
	\renewcommand{\arraystretch}{1.3}
	\begin{center}
	\begin{tabular}{l|c|c|c||c|c}
		& \cite{xu2017scene} & VGfM-2D & Geometric encoder & VGfM &  VGfM +Fusion\\
		\hline
		Object label&	74\%	 &   74\%  & 58\% &  75\% & \textbf{76}\%  \\
		\hline
		\hline
		Same-plane   &	74\% &   74\%  & 70\% &  74\% & \textbf{78}\%  \\
		Same-set     &	58\%	 &   59\%  & 55\% &  61\% & \textbf{62}\%  \\
		Support      &	62\%	 &   64\%  & \textbf{85}\%  & 67\% & 64\%  \\
		Part-of      &	68\%	 &   69\%  & \textbf{80}\%  & 69\% & 59\%  \\ 
		\hline
	\end{tabular}
	\end{center}
	\caption{This table shows the accuracy for the prediction of each predicate and the  object labels. The numbers in bold are the best performing methods.}
	\label{tab:predicate}
\end{table}
The results of the baseline method do not exceed $75\%$ of accuracy, which suggests that this task is difficult especially for the high level \textit{Same-set} predicate. 
As shown on the second column, augmenting the appearance with the 2D coordinates allows VGfM-2D to obtain an improvement of $1-2\%$ as it is commonly observed in computer vision for this kind of feature augmentation.

The results of the geometric encoder shows large differences between the tasks. The low performance for the task of object classification is not surprising as a 3D bounding box alone carries little information about the object label. Regarding the results for the predicate prediction, we tested the same architecture but providing the GT ellipsoids and found only a difference of $1-5\%$ depending on the predicates. It is thus possible that some errors are due to partly segmented objects in the annotations, resulting in inaccurate bounding boxes. We also observe that results are higher than any other method for the predicates \textit{Support} and \textit{Part-of}. One explanation is that these two predicates are less frequent in the dataset (respectively 3000 and 600 instances compared to around 30000 for the other classes). In these cases with less training data, having a more simple, shallower architecture with a reduced number of parameters helps. Unfortunately, standard data augmentation techniques such as cropping or shifting cannot be directly applied to augment the number of samples as they would introduce incoherences with the 3D geometry.

The proposed single image VGfM method has a better or similar accuracy than the methods which do not use 3D information. This suggests that the information contained in the ellipsoids is beneficial for predicting relationships and that our model is able to use it. 
We can draw similar conclusions for the fusion mechanism. Indeed the fusion mechanism shows improvements for the predicates which are common on the dataset. For these cases, the model successfully manages to leverage on the different sources of information to reach an improvement of accuracy of $4\%$ with respect to the initial baseline. However, it fails to improve for the ones which contain only a few training examples. This effect should be more important for the fusion mechanism since in this case, one training sample corresponds to a sequence of 10 images. Thus the number of instances in the training data is roughly divided by 10.

\begin{figure}[htbp!]  	
	\centering
	\includegraphics[width=\linewidth]{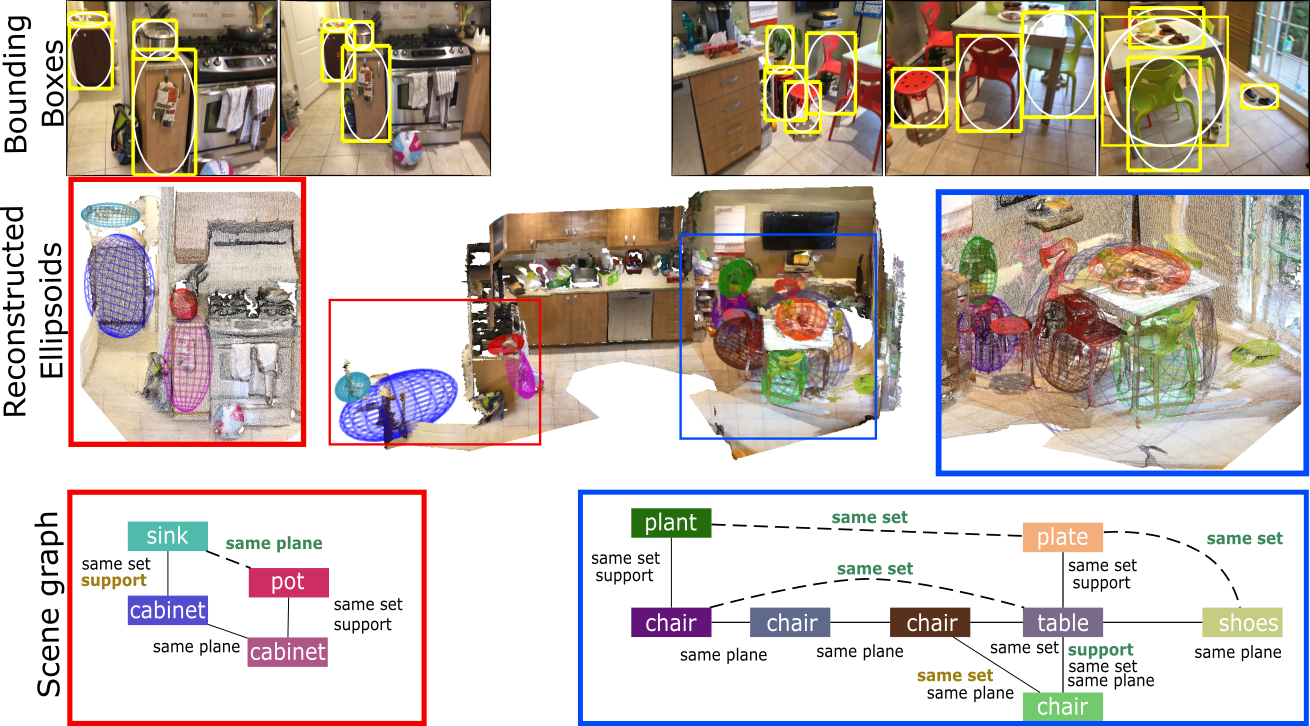} 
	\caption{The top row shows images extracted from two sequences together with the bounding box detections in yellow and in white the conics used to estimate the ellipsoids. The second row shows the resulting ellipsoids of these two sequences as well as the global object layout of the room. The third row shows the corresponding scene graphs obtained with our proposed approach. We did not display all the relations to ease the visualization. Predicates in bold brown font are miss-detections and bold green font with dashed line are false alarms (best viewed in color).}
	\label{fig:qual}
\end{figure}

Fig.~\ref{fig:qual} shows some qualitative results of two image sequences coming from the same room. 
On the left, the model successfully identified the two sets of  objects, and it detects that the two cabinets are on the same plane. Since perspective effects are strong, this reasoning would be difficult with 2D features only. The right part is a complex scene with many overlapping objects. Although some errors are still present, leveraging over multiple views provides, as a 3D graph, a rich description of the scene which could enable further high level reasoning.

\section{Conclusions} \label{sec:conc}
We addressed the problem of generating a 3D scene graph from multiple views of a scene. The VGfM approach leverages both geometry and visual appearance and it learns to refine globally the features  and to merge the different sources of information through a message passing framework. We have evaluated on a new dataset which focuses on the relationships in 3D and show that our method outperforms a 2D baseline method. 

The problem of creating a scene graph in both 2D and 3D from multiple views has been addressed for the first time in this paper, however there are many areas to be explored that can enhance  performances. First, other sources of knowledge could be used. In particular, \cite{zellers2017neural} shows that the manifold of the scene-graph is rather low dimensional as many of them contain recurrent patterns. This suggests that a strong prior could be built to encode this topology. Secondly, the knowledge about the visual appearance and the semantic relationships could be used to refine the geometric nodes by refining the quality of the ellipsoids. Last but not least, the case of dynamic scene could be investigated. As the predictions of our model are done per image, it can be readily applied on this setting.

\bibliographystyle{splncs04}
\bibliography{egbib}

\newpage

\appendix

\begin{adjustwidth}{-100pt}{-100pt}
\begin{center}
\LARGE Supplementary Material: Mathematical details on the formulation of LfD
\end{center}

~

~

This supplementary material provides additional details about the derivation of the LfDC method outlined in Sec. 3 of the submitted paper. We first describe how to derive Eq. (1) used by the LfD method i.e. how the estimation of the quadrics can be linearised in the dual space. Then we provide the full expression of the dual quadrics and conics to justify Eq. (7). 

\section{LfD approach: from primal to dual space}

As stated in the paper, we are interested in finding the 3D ellipsoids $\mat Q_i$ whose projections onto the image planes best fit the 2D ellipses $\hat{\mat C}_{if}$. Each ellipse $\hat{\mat C}_{if}$ correspond to the detection of the object $i$ in the image $f$, with $i=1 \dots N$ and $f=1 \dots F$. The relationship between $\mat{Q}_i$ and their reprojected conics $\mat{C}_{if}$ is defined by the $3\times4$ perspective camera matrices $\mat P_{f}$ which are assumed to be known. The projection of conics from 3D to 2D is a complex operation in the primal space however this operation is highly simplified in the dual space. 
We will now describe the steps which show that the problem can be linearised. 

The dual quadric is defined by the matrix  $\mat Q^\circ_i = adj(\mat Q_i)$, where $adj$ is the adjoint operator, and the dual conic is defined by ${\mat{C}}^\circ_{if} = adj({\mat C}_{if})$ \cite{hartley2003multiple}. Considering that the dual conic ${\mat C}^\circ_{if}$, like the primal one, is defined up to an overall scale factor $\beta_{if}$, the relation between a dual quadric and its dual conic projections ${\mat C}^\circ_{if}$ can be written as:

\begin{equation} \label{eq:dualq2c}
\beta_{if} {\mat{C}}^\circ_{if} =  \mat P_{f}  \mat Q^\circ_i  \mat P^{\top}_{f}.
\end{equation}

In order to recover $\mat Q^{\circ}_{i}$ in closed form from the set of dual conics $ \lbrace {\mat C}^\circ_{if} \rbrace_{f=1 \dots F} $, Eq. (\ref{eq:dualq2c}) is re-arranged into a linear system. Let us define  $\vec v^{\circ}_{i} = vech (\mat Q_i^\circ)$ and ${\vec c}^{\circ}_{if} =vech ({\mat C}^\circ_{if})$ as the vectorisation of symmetric matrices $\mat Q^{\circ}_{i}$ and ${\mat C}^\circ_{if}$ respectively\footnote{The operator $vech$ serialises the elements of the lower triangular part of a symmetric matrix, such that, given a symmetric matrix $\mat X \in \mathbb{R}^{n \times n}$, the vector $\vec x$, defined as $\vec x = vech(\mat X)$, is $\vec x \in \mathbb{R}^{g} $ with $ g=\frac{n(n+1)}{2}$.}. Then, let us arrange the products of the elements of $\mat P_{f}$ and $\mat P^{\top}_{f}$ in a unique matrix $\mat G_{f} \in \mathbb R^{6 \times 10}$ as follow \cite{henderson1979vec}:
\begin{equation}
\mat G_f  = \mat D (\mat P \otimes \mat P) \mat E,
\end{equation}
where $\otimes$ is the Kronecker product and matrices $\mat D \in \mathbb{R}^{6 \times 9}$ and $\mat E \in \mathbb{R}^{16 \times 10}$ are two matrices such that $vech (\mat X) =\mat D \hspace{0.1 cm} vec(\mat X)$ and $vec (\mat Y) =\mat E \hspace{0.1 cm} vech(\mat Y)$ respectively,  where $\mat X \in \mathbb{R}^{9 \times 9}$ and $\mat Y \in \mathbb{R}^{16 \times 16}$ are two symmetric matrices\footnote{The operator $vec$ serialises all the elements of a generic matrix.}. 
This matrix is defined as:
	\begin{equation} 
		\arraycolsep=1.7pt
		\scriptsize
		\mat G_f \scalebox{0.5}[1.0]{$=$} 
		\left[ \begin {array}{cccccccccc} {p_{{11}}}^{2}&2\,p_{{12}}p_{{11}}&
		2\,p_{{13}}p_{{11}}&2\,p_{{14}}p_{{11}}&{p_{{12}}}^{2}&2\,p_{{13}}p_{{
				12}}&2\,p_{{14}}p_{{12}}&{p_{{13}}}^{2}&2\,p_{{13}}p_{{14}}&{p_{{14}}}
		^{2}\\ \noalign{\medskip}p_{{21}}p_{{11}}&p_{{21}}p_{{12}}+p_{{22}}p_{
			{11}}&p_{{23}}p_{{11}}+p_{{21}}p_{{13}}&p_{{24}}p_{{11}}+p_{{21}}p_{{
				14}}&p_{{22}}p_{{12}}&p_{{22}}p_{{13}}+p_{{23}}p_{{12}}&p_{{22}}p_{{14
			}}+p_{{24}}p_{{12}}&p_{{23}}p_{{13}}&p_{{23}}p_{{14}}+p_{{24}}p_{{13}}
			&p_{{24}}p_{{14}}\\ \noalign{\medskip}p_{{31}}p_{{11}}&p_{{31}}p_{{12}
			}+p_{{32}}p_{{11}}&p_{{33}}p_{{11}}+p_{{31}}p_{{13}}&p_{{34}}p_{{11}}+
			p_{{31}}p_{{14}}&p_{{32}}p_{{12}}&p_{{32}}p_{{13}}+p_{{33}}p_{{12}}&p_
			{{32}}p_{{14}}+p_{{34}}p_{{12}}&p_{{33}}p_{{13}}&p_{{33}}p_{{14}}+p_{{
					34}}p_{{13}}&p_{{34}}p_{{14}}\\ \noalign{\medskip}{p_{{21}}}^{2}&2\,p_
			{{22}}p_{{21}}&2\,p_{{23}}p_{{21}}&2\,p_{{24}}p_{{21}}&{p_{{22}}}^{2}&
			2\,p_{{23}}p_{{22}}&2\,p_{{24}}p_{{22}}&{p_{{23}}}^{2}&2\,p_{{23}}p_{{
					24}}&{p_{{24}}}^{2}\\ \noalign{\medskip}p_{{31}}p_{{21}}&p_{{31}}p_{{
					22}}+p_{{32}}p_{{21}}&p_{{33}}p_{{21}}+p_{{31}}p_{{23}}&p_{{34}}p_{{21
				}}+p_{{31}}p_{{24}}&p_{{32}}p_{{22}}&p_{{32}}p_{{23}}+p_{{33}}p_{{22}}
				&p_{{32}}p_{{24}}+p_{{34}}p_{{22}}&p_{{33}}p_{{23}}&p_{{33}}p_{{24}}+p
				_{{34}}p_{{23}}&p_{{34}}p_{{24}}\\ \noalign{\medskip}{p_{{31}}}^{2}&2
				\,p_{{32}}p_{{31}}&2\,p_{{33}}p_{{31}}&2\,p_{{34}}p_{{31}}&{p_{{32}}}^
				{2}&2\,p_{{33}}p_{{32}}&2\,p_{{34}}p_{{32}}&{p_{{33}}}^{2}&2\,p_{{33}}
				p_{{34}}&{p_{{34}}}^{2}\end {array} \right]
				\label{eq:mxG}
			\end{equation}
 %
Given $\mat G_{f}$, we can rewrite Eq. (\ref{eq:dualq2c}) as~\cite{rubino20173d}:
\begin{equation} \label{eq:eqOut}
\beta_{if} {\vec c}^\circ_{if} = \mat G_{f} \vec v^{\circ}_{i},
\end{equation}
which corresponds to the Eq. (1) used in our ACCV 2018 submission. 

\section{Derivation of the constraints on the conics/quadrics centers in LfDC}

\subsection{Selecting conic translational components}

We first consider the expression of the dual conic $\mat C_{if}^\circ$. Every conic can be centered in the image center with normalised axes length using the $3 \times 3$ transformation  matrix $\mat H_{if}$ as follow:  
\begin{equation}
\label{eq:centering}
\mat C_{if}^\circ = \mat H_{if}  \breve{\mat{C}}_{if}^\circ  \mat H_{if}^{\top},
\end{equation}
with:
\begin{equation}\label{eq:matHC}
\mat H_{if} \! = \!\! 
\begin{bmatrix}
h & 0   & t^c_1 \\
0   & h & t^c_2 \\
0  &  0   & 1 
\end{bmatrix},
\hspace{0.2cm}
\breve{\mat{C}}_{if}^\circ \! = \!  
\begin{bmatrix}
c^\circ_{11} & c^\circ_{12}   & 0 \\
c^\circ_{12}   & c^\circ_{22} & 0 \\
0  &  0   &  \minus 1 \\
\end{bmatrix},
\end{equation}
where $t^c_1$ and $t^c_2$  are the coordinates of the ellipse center and $h=\sqrt{l_1^2 + l_2^2}$, where $l_1,l_2 \in \mathbb{R}$ are the two semi axes of the ellipse.
Using Eqs. (\ref{eq:centering}) and (\ref{eq:matHC})
we can express the vectorised conic $\vec c_{if}^\circ$ as: 
\begin{equation}
\vec c^\circ_{if} =  
\left[ 
\begin {array}{c}
\noalign{\medskip} {h}^{2}c^\circ_{11}-{t^c_1}^{2} \\
\noalign{\medskip} {h}^{2}c^\circ_{12} - t^c_1 t^c_2\\ 
\noalign{\medskip} -t^c_1 \\
\noalign{\medskip} {h}^{2}c^\circ_{{22}}-{t^c_2}^{2} \\ 
\noalign{\medskip} -t^c_2\\ 
\noalign{\medskip} -1
\end {array} 
\right].
\end{equation}
We can see that the third and fifth components of the vector, i.e. $\vec c^*_{ij} = \vec c^\circ_{if (3,5)}$, contain the two translation parameters as stated in the Eq. (5) of the ACCV 2018 submission.

\subsection{Selecting quadric translational components}

The same procedure can be applied to the quadrics  $\mat Q_i^\circ$ in dual space but this time defining the $4 \times 4$ matrix $\mat Z$ giving: 
\begin{equation}
\label{eq:quadric_norm}
\mat Q_i^\circ = \mat Z \breve{\mat {Q_i}}^\circ \mat Z^\top
\end{equation}
where 
$\breve{\mat {Q}}^\circ$ is an ellipsoid centered on the origin and with the axes aligned to the 3D coordinates and $\mat Z$ is an homogeneous transformation accounting for an arbitrary rotation and translation. The matrices $\mat Z$ and $\breve{\mat {Q}}^\circ$ can be written respectively as: 
\begin{equation} \label{eq:transQuad}
\mat Z =
\begin{bmatrix}
\mat R(\vec \theta) & \vec t \\
\vec 0_3^{{\top}} 	& 1 \\
\end{bmatrix},
\,\,\,
\breve{\mat {Q}_i}^\circ = 
\begin{bmatrix}
a^2 & 0 & 0 & 0 \\
0 & b^2 & 0 & 0 \\
0 & 0 & c^2 & 0 \\ 
0 & 0 & 0 & \minus 1 \\
\end{bmatrix}
\end{equation}
where $\vec t = [t_1,t_2,t_3]^{{\top}}$ is the 3D translation vector, $\mat R (\vec{\theta})$ is the rotation matrix function of the Euler angles $\vec{\theta} = [\theta_1, \theta_2,\theta_3]^{{\top}}$ and $a, b, c$ are the three semiaxes of the ellipsoid\footnote{The positivity of $a^2,b^2,c^2$ grants that $\mat L^\circ$ represents an ellipsoid and not a generic quadric.}. Therefore, we can express every ellipsoid in terms of the nine parameters  $ \theta_1, \theta_2, \theta_3, t_1, t_2, t_3, a, b, c$. Now, defining the vector  $\vec e \in \mathbb{R}^{9+F}$ as $\vec e = [\theta_1, \theta_2, \theta_3, t_1, t_2, t_3, a, b, c, \beta_{1}, \dots \beta_{F}]^{\top}$ we can evaluate a functional form of the vector $\vec{v}_i^\circ(\vec e)$ as follow:

\begin{equation} \label{eq:vectorwofe}
\scriptsize
\vec{v}_i^\circ(\vec e) = 
\left[ \begin {array}{c}  r_{11}(\vec \theta)^{2}{\it a^2} + {r_{{12}}}(\vec \theta)^{2}{\it b^2} + r_{13}(\vec \theta)^{2}{\it c^2} - t_{1}^{2}\\ 
\noalign{\medskip}  r_{{11}}(\vec \theta) r_{21}(\vec \theta) {\it a^2}+ r_{12}(\vec \theta) r_{{22}(\vec \theta)}{\it b^2}+ r_{{13}}(\vec \theta) r_{{23}}(\vec \theta){\it c^2} - t_{{1}}t_{{2}}\\ 
\noalign{\medskip} r_{{11}}(\vec \theta) r_{{31}}(\vec \theta) {\it a^2}+ r_{12}(\vec \theta) r_{32}(\vec \theta){\it b^2}+ r_{{13}}(\vec \theta) r_{{33}}(\vec \theta) {\it c^2} - t_{{1}}t_{{3}} \\ 
\noalign{\medskip}-t_{{1}}\\ 
\noalign{\medskip} {r_{{21}}}(\vec \theta)^{2}{\it a^2}+ {r_{{22}}}(\vec \theta)^{2}{\it b^2}+ r_{23}(\vec \theta)^{2}{\it c^2} - t_{2}^{2}\\ 
\noalign{\medskip} r_{21}(\vec \theta)r_{31}(\vec \theta) {\it a^2}+ r_{22}(\vec \theta) r_{32}(\vec \theta){\it b^2}+ r_{{23}}(\vec \theta)r_{33}(\vec \theta){\it c^2}- t_{{2}}t_{{3}}\\
\noalign{\medskip}-t_{{2}}\\ 
\noalign{\medskip} {r_{{31}}}(\vec \theta)^{2}{\it a^2}+ {r_{{32}}}(\vec \theta)^{2}{\it b^2}+ {r_{{33}}}(\vec \theta)^{2}{\it c^2}- {t_{{3}}}^{2}\\ 
\noalign{\medskip}-t_{{3}}\\ 
\noalign{\medskip} -1 \\ 
\noalign{\medskip} \beta_1 \\
\noalign{\medskip} \vdots \\
\noalign{\medskip} \beta_F  \end {array}
\right]
\end{equation} 
where the terms in $r_{mn} \,\,| \,\, m,n = 1, \dots, 3$ are the entries of the rotation matrix $\mat R(\vec \theta)$.
We can see again that the fourth, seventh and ninth components of the vector contain the translation paramater as defined by  Eq. (5), i.e. $\vec v^* = \vec{v}^\circ_{i(4,7,9)}$, of the ACCV 2018 submission.

\subsection{Selecting matrix $\mat G_f$ components}

Similarly, we can extract the components of the $6 \times 10$ matrix $\mat G_f$  related to the centers of both conics and quadrics. Before that, notice that if we apply the centering and normalisation as given in Eq. (\ref{eq:quadric_norm}) this has an effect on the projection relation in Eq. \ref{eq:dualq2c}. In practice, we obtain that the form of the matrix $\mat G_f$ given the centering is now given by Eq. (\ref{eq:mxG}) \cite{Crocco:etal:2016}. Now,  we need to select the  components related to the translational elements of the conics and quadrics, i.e. the rows (3, 5). In practice, this results in selecting specific elements of the projective camera matrix $\mat P_i$.

In order to be consistent with the solution of the linear system (which is the objective function of LfD): 
\begin{equation}\label{eq:eqsys}
\mat M_{i} \vec w_{i} = \vec 0_{6F},
\end{equation}
where, 
\begin{equation}\label{eq:eqsysmx}
\mat M_{i} = 
\arraycolsep=1.4pt
\begin{bmatrix}
\mat G_1 	& \minus {\vec c}_{i1}	& \vec 0_6 		& \vec 0_6 		& \ldots & \vec 0_6 \\
\mat G_2 	& \vec 0_6   	& \minus {\vec c}_{i2}	& \vec 0_6  		& \ldots & \vec 0_6 \\
\vdots 		&   	&  		&    		& \ddots &  \\
\mat G_F	& \vec 0_6  	& \vec 0_6 		& \vec 0_6   		& \ldots & \minus {\vec c}_{iF} \\
\end{bmatrix}, \;\;\;
\vec w_{i} = 
\begin{bmatrix}
\vec v_i\\
\vec{\beta}_i
\end{bmatrix},
\end{equation}

we have that the final $2 \times 10$ matrix $\mat G^c_f$ is given by:

\begin{equation} 
		\arraycolsep=1.7pt
		\mat G^c_f = \breve{\mat G}_{f(3,5)\times (1:10)} =
		\left[ \begin {array}{cccccccccc} 0 & 0 & 0 & p_{{11}} & 0 & 0 &p_{{12}} &0 & p_{{13}}& p_{{14}}\\ 
		\noalign{\medskip}0 & 0 & 0 & p_{{21}} & 0 & 0 &p_{{22}} &0 & p_{{23}}& p_{{24}}\end {array} \right].
\end{equation}

The elements $\mat G^c_f$, $\vec c^*$, $\vec v^*$ can be now plugged into the matrix $\tilde{\mat M}_{i}$ giving:

\begin{equation}\label{eq:eqsysmx_cen}
\tilde{\mat M}_{i} = 
\arraycolsep=1.4pt
\begin{bmatrix}
\mat G_1 	& \minus {\vec c}_{i1} 	& \vec 0_6 		& \vec 0_6 		& \ldots & \vec 0_6 \\
\mat G^c_1 	& \minus {\vec c}_{i1}^* 	& \vec 0_6 		& \vec 0_6 		& \ldots & \vec 0_6 \\
\vdots 		&   	&  		&    		& \ddots &  \\
\mat G_F	& \vec 0_6  	& \vec 0_6 		& \vec 0_6   		& \ldots & \minus {\vec c}_{iF} \\
\mat G^c_F	& \vec 0_6  	& \vec 0_6 		& \vec 0_6   		& \ldots & \minus c^*_{iF} \\
\end{bmatrix}.
\end{equation}
The solution of the linear system:
\begin{equation} \label{eq:argmin}
 \tilde{\vec{w}}_{i}= \arg \min_{\vec w} \| \tilde{\mat M}_{i} \vec w \|_2^2 \hspace{0.3cm} s.t. \hspace{0.2cm} \|\vec w\|^2_2=1,
\end{equation}
provides the solution of the LfDC problem.

\begin{table*}[t!]
\begin{equation}
\arraycolsep=1.7pt
\scriptsize
\breve{\mat G}_f =
 \left[ \begin {array}{cccccccccc} {p_{{11}}}^{2}&2\,p_{{12}}p_{{11}}&
2\,p_{{13}}p_{{11}}&2\,p_{{14}}p_{{11}}&{p_{{12}}}^{2}&2\,p_{{13}}p_{{
12}}&2\,p_{{14}}p_{{12}}&{p_{{13}}}^{2}&2\,p_{{13}}p_{{14}}&{p_{{14}}}
^{2}\\ \noalign{\medskip}p_{{21}}p_{{11}}&p_{{21}}p_{{12}}+p_{{22}}p_{
{11}}&p_{{23}}p_{{11}}+p_{{21}}p_{{13}}&p_{{24}}p_{{11}}+p_{{21}}p_{{
14}}&p_{{22}}p_{{12}}&p_{{22}}p_{{13}}+p_{{23}}p_{{12}}&p_{{22}}p_{{14
}}+p_{{24}}p_{{12}}&p_{{23}}p_{{13}}&p_{{23}}p_{{14}}+p_{{24}}p_{{13}}
&p_{{24}}p_{{14}}\\ \noalign{\medskip}0&0&0&p_{{11}}&0&0&p_{{12}}&0&p_{{13}}&p_{{14}}\\ \noalign{\medskip}{p_{{21}}}^{2}&2\,p_
{{22}}p_{{21}}&2\,p_{{23}}p_{{21}}&2\,p_{{24}}p_{{21}}&{p_{{22}}}^{2}&
2\,p_{{23}}p_{{22}}&2\,p_{{24}}p_{{22}}&{p_{{23}}}^{2}&2\,p_{{23}}p_{{
24}}&{p_{{24}}}^{2}\\ \noalign{\medskip}0&0&0&p_{{21
}}&0&0
&p_{{22}}&0&p_{{23}}&p_{{24}}\\ \noalign{\medskip}0&0&0&0&0&0&0&0&0&1\end {array} \right]
\label{eq:mxG}
\end{equation}
\vspace{-.6cm}
\end{table*}
\end{adjustwidth}

\end{document}